# Viewing Robot Navigation in Human Environment as a Cooperative Activity

Harmish Khambhaita and Rachid Alami

**Abstract** We claim that navigation in human environments can be viewed as cooperative activity especially in constrained situations. Humans concurrently aid and comply with each other while moving in a shared space. Cooperation helps pedestrians to efficiently reach their own goals and respect conventions such as the personal space of others. To meet human comparable efficiency, a robot needs to predict the human trajectories and plan its own trajectory correspondingly in the same shared space. In this work, we present a navigation planner that is able to plan such cooperative trajectories, simultaneously enforcing the robot's kinematic constraints and avoiding other non-human dynamic obstacles. Using robust social constraints of projected time to a possible future collision, compatibility of human-robot motion direction, and proxemics, our planner is able to replicate human-like navigation behavior not only in open spaces but also in confined areas. Besides adapting the robot trajectory, the planner is also able to proactively propose co-navigation solutions by jointly computing human and robot trajectories within the same optimization framework. We demonstrate richness and performance of the cooperative planner with simulated and real world experiments on multiple interactive navigation scenarios.

## 1 INTRODUCTION

Taking inspiration from the *joint action* literature [1] and from our previous contributions on robot planning abilities for human-robot task achievement [2], we propose a reactive navigation planner that builds and maintains a set of streams of execution for the robot and the humans in its close vicinity. Indeed, it has been shown

Harmish Khambhaita
LAAS-CNRS, Universit de Toulouse, CNRS, Toulouse, France, e-mail: `harmish@laas.fr`

Rachid Alami
LAAS-CNRS, Universit de Toulouse, CNRS, Toulouse, France, e-mail: `alami@laas.fr`

that it is sometimes pertinent to endow the robot with the ability to plan not only for itself but also for its human partner. This ability takes its full meaning and pertinence when it is necessary that both act in order to solve a problem. In navigation, this corresponds to very constrained environments.

While navigation in a populated environment can generally be modeled as a kind of coordination activity, since each individual has his own goal and all share an environment, this very same activity can be transformed into a problem that needs cooperation of two or several individuals when the environment becomes very constrained: a given individual cannot find his path unless another individual participates and helps in finding a solution.

This is exactly the kind of problems we want to tackle:

- We would like to develop a robot navigation system that is able to manage usual coordination issues, but that is also able to manage intricate situations.
- Besides, we would like, to come up with a scheme that allows the robot to be proactive by proposing an acceptable solution and, whenever possible, to take "most of the load" when the human and the robot have to share the load to solve a problem.
- And, finally, we would like the robot to take into account human acceptability and comfort issues.

Fig. 1 shows a typical case where both robot and human can safely and smoothly pass each other if they cooperate and facilitate the other party by giving enough space to move. It is the duty of both to avoid a collision and help the other party to advance towards their destination. The solution does not include only the contribution of the robot but also of the human. This is why we claim that our planner is a cooperative planner.

In this work, we propose a cooperative navigation planner that predicts a plausible trajectory for the humans and accordingly plans for a robot trajectory that satisfies a set of social constraints. It generates both robot and human trajectories within a unified planning framework, thus facilitating both agents to avoid any other static or dynamic obstacle present in the shared space. Generation of these trajectories becomes one multi-constrained problem and it is solved using a graph-based optimal solver. We not only use proxemics, but also apply *time-to-collision* and *directional* constraints during optimization. Another important aspect in terms of perceived safety and comfort is that the proposed planner inherently balances between trajectory modification and speed adaptation. We also show improvements in the fluency of interaction with use of the proposed cooperative navigation planner.

The key contributions of this paper are three-fold:

1. An optimization based framework for computing the robot trajectory and predicting probable trajectories of nearby humans that respect motion and social constraints.
2. Prudently devised social constraints for (a) safety, (b) time-to-collision, and (c) directional compatibility of human-robot motion.
3. A demonstration of the success of proposed approach in everyday interactive navigation situations, especially in confined spaces.

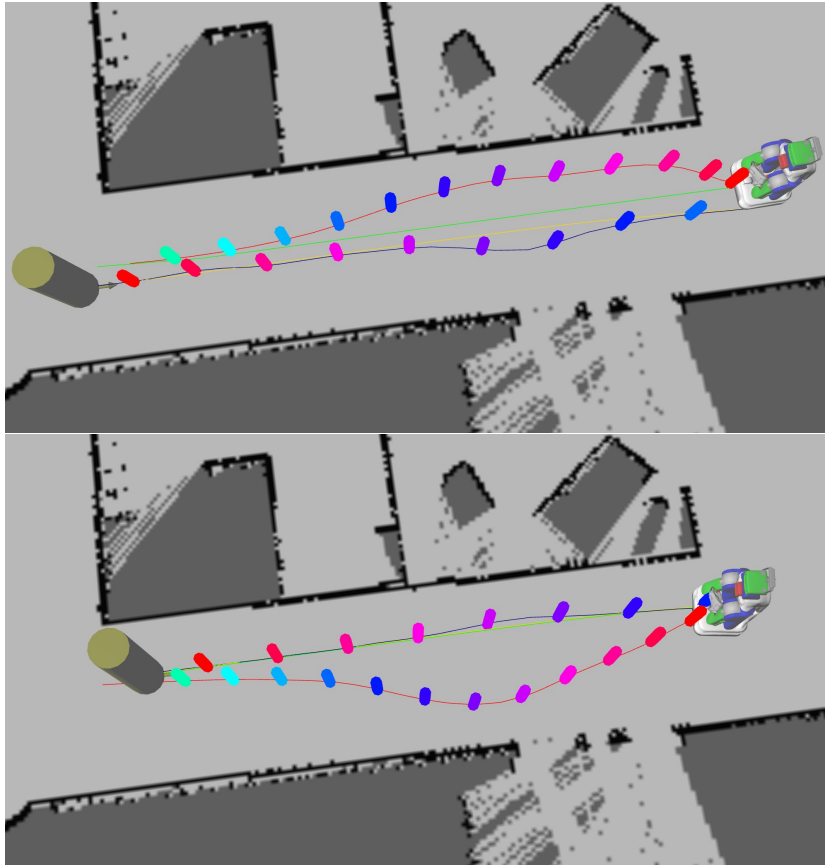

Fig. 1: A corridor-crossing scenario with proposed cooperative planner. The robot is able to calculate its own trajectory (in red) and proposes a trajectory for the human (in blue) that solves the co-navigation problem. Here the robot assumes that human wants to go to the other end of the corridor. We have added cylindrical shaped landmarks on the planned robot trajectory and predicted human trajectory by the cooperative planner. The landmarks shows future human and robot positions at every second. The color of the landmark on the human and robot trajectories correspond to same future time. We can see that the robot moves to its right well in advance (even if it is not absolutely necessary at the moment the robot sees the human). With such proactive behavior the robot is offering the solution to the co-navigation situation. The robot does not accelerate until the human passes so that the human do not feel threatened, in this example it passes the human at approximately $0.6 \,\text{m/s}$ and later accelerates to $0.8 \,\text{m/s}$. If the human reject the solution (bottom figure) suggested by the robot and move in the opposite direction to what the robot has predicted, the robot is able to quickly react by changing its path that adheres to human's wish. Robot's global path is shown in green, where as a velocity based prediction of human path is shown in yellow.

Our approach aims to balance and tune the efforts between the human and the robot to solve a co-navigation task. It provides additional advantage of streamlining robot behavior from open spaces to very constrained environmental conditions. In its spirit, our approach is similar to the previously proposed approaches for geometric [3] and symbolic [4] planning systems, where the robot synthesizes a shared plan for the human and itself.

## 2 RELATED WORKS

### *Acting together*

Research in psychology [5] and philosophy [6] have led to a good understanding of human behavior during joint actions and collaboration and have helped to identify the key elements for human-robot joint actions [7].

In [8], Tomassello et al. define a goal as the representation of a desired state and an intention as an action plan chosen in order to reach the goal. Bratman adds that if there is a *shared intention* to perform an action, the partners should agree on the meshing sub-parts of a *shared plan* [9], which is elaborated based on *common ground* [10].

The reactive scheme that we propose here implements some of the joint action principles. It takes into account the (navigation) intention of the human; it adapts permanently to his behavior; it is proactive and does its best to facilitate the action of the human.

### *Social conventions*

The theory of proxemics [11] has provided rules for realizing more human like behavior during robot motion and non-motion tasks [12]. The most of state-of-the-art human-aware navigation planners add proxemics costs around humans in a grid-based map representation of the robot operating environment [13]. Using this *costmap* the path planning algorithms can generate paths that lower the total cost over the entire path, thereby keeping a safe distance from humans to maximize human comfort. The human-aware navigation planner described by Sisbot et al. [14] already provide safe paths considering not only proxemics distance but also other social criteria like visibility and hidden zones around static humans in the environment.

In extension to that work we define a safety constraint in optimization framework around the trajectories of moving humans. However, to avoid over-cautious behavior of the robot, we only apply the safety costs around the respective points in time along the human-robot trajectory pairs. That means, the optimization procedure ensures given safety distance between planned position of the robot and predicted position of the human at all future time points $t = 1, t = 2, \ldots$ up to the planning horizon.

Since these cost generating functions regard humans as static obstacles, typically the planning frameworks perform continuous re-planning of the robot path to cope up with dynamic situations. Nevertheless, only immediate path of the robot is re-planned for the planning algorithm to reach real-time compliance. The resulting robot motions are robust and safe but not necessarily social. The robot often oscillates or stops completely while moving near humans [15].

The directional cost model introduced by Kruse et al. [16] have shown to increase legibility of the robot motions, where a robot attempts to solve a spatial conflict by adjusting velocity instead of path when possible. Humans prefer robot following this strategy, particularly in path crossing situations [15]. We exploit this result and introduce *directional costs* in our optimization framework. The *directional costs* discourages face-to-face motion towards a person. It also makes the robot slow down while moving very near to the human because of the modeled inverse proportionality of this constraint to the distance between the human and the robot.

### *Human motion prediction*

The requirement for human motion prediction arises when we intend to design a robot navigation system that is socially acceptable [17]. Prediction of human trajectories independent of robot plans, however, does not alleviate the problem of purely reactive robot behavior [18]. For example, consider a corridor situation where a robot and a person could only cross each other in a side-by-side configuration (fig. 1). If the person is walking in the middle of the corridor, due to their predicted path, the robot will fail to find a collision free trajectory using a reactive planner. Therefore, prediction of human trajectories have to consider that the humans do see the robot and will also try to contribute to collision avoidance with the robot by modifying their own trajectories. In other words, there is a need for a planner that can proactively suggest a solution to the interactive navigation situations.

More recent approaches have paid scrupulous attention to human path prediction based on renowned social force model [19], which facilitates the navigation planner to cope with uncertain human motions. These methods predict a class of homotypically distinct trajectories for humans and design a planner that, from human demonstrations, can learn navigation policies for robot to move on human-like trajectories [17]. Although, this scheme works fine in large or open spaces where the robot have enough latitude to move, it may require re-learning of the model parameters for specialized or constrained situations, such as crossing long corridors or passing through a door.

*Planning for the robot and the human*

Concerning the ability for the robot to plan not only for itself but also for its human partner, we have developed earlier a task planner called HATP planner (*Human Aware Task Planner*) [20, 21]. The HATP planning framework extends the traditional hierarchical task network planning domain representation and semantics by making them more suitable to produce plans which involve humans and robots acting together toward a joint goal. HATP is used by the robot to produce human-robot *shared plans* which are then used to anticipate human action, to suggest a course of action to humans, or possibly to ask help from the human if needed.

This effectively enriches the interaction capabilities of the robot by providing the system with what is in essence a prediction of the human behavior. This prediction is also used by the robot execution controller to monitor the engagement of the human partner during plan achievement. Another key property is to produce plans that would be possibly preferred by the human partner. For instance, HATP includes cost-based plan selection as well as mechanisms called *social rules* to promote plans that are considered as suitable for human-robot interaction.

We have also applied the same approach to a different type of problem that calls for elaborate geometric reasoning and planning abilities: robot-human handover in a workspace possibly cluttered by obstacles. The question is where to perform the task and how to balance between the efforts of the human and the robot [22, 3].

Similarly, the planner we propose here manages explicitly one elastic band per agent and plans for all. A number of social constraints have been specially devised to produce plans that would be possibly preferred by the human encountering the robot. The robot and the human bands tightness are different in order to force the robot take most of the load.

Another approach, presented by Ferrer et al. [23], uses the social force model for both to predict human paths and control the robot motion. In this approach, every iteration of planning step uses the human prediction information which is dependent on the path calculated during the previous iteration. Advantage of such scheme is robot acting proactively in given situation, however, human prediction is only reaction to the robot motion. Our approach is rather cooperative, where optimization process coherently provides a solution for the interactive navigation situation. Comparable results could be achieved with proposed cooperative planner in non-constrained cases, but in situations such as door crossing the cooperative planner avoids unnecessary detours. In situations like corridor crossing with cooperative planner the robot prefers waiting in a place where it limits, as much as possible, obstruction to the human motion, instead of moving backwards due to repulsive human interaction forces when using a social force model based planner.

The intention aware reactive avoidance scheme proposed by [24] uses counterfactual reasoning to calculate probabilities over a possible set of navigational goals. Using such probability set, this approach predicts human motion towards most probable goal and generates locally optimal motions for multiple robots. The time scaled collision cone based approach aims to solve the same problem, albeit giving same treatment to human and non-human obstacles [25]. While being effective in densely crowded environments, as a virtue of remaining purely reactive, such approaches could lead to needless detours in intricate situations. Our focus is, rather, on sophisticated interactive motion with single person to a group of people in semi crowded environments.

## 3 METHODOLOGY

Elastic band is a well-studied approach for dynamic obstacle avoidance that only locally modifies the robot path to keep a safe distance from previously unknown obstacles [26]. However, the modified path often does not satisfy the kinodynamic constraints of the robot. Therefore, a general scheme is to use a controller module that takes the output path (elastic band) and generates feasible trajectories that the robot can follow [27]. Recent proposal of *timed elastic band* evades this problem by explicitly considering temporal information [28]. It locally deforms the robot path and computes a trajectory augmented with a series of time-difference values between each successive poses, instead of a purely geometric path. *Timed elastic band* makes it easy to take kinodynamic and nonholonomic constraints into account, formalizing the optimization problems as a non-linear least squares problem. We have substantially extended this work by introducing prediction and optimization of human trajectories in the same framework, in [29] we gave preliminary description of out approach. Besides, we have brought in carefully selected social constraints in to the optimization framework.

We address the task of robot navigation among humans by applying least squares optimization to simultaneously minimize multiple cost-functions that represent costs associated with human-robot cooperation as well as robot dynamics. The optimization framework explicitly includes prediction of plausible trajectories within the same non-linear least squares problem. Compared to earlier work [30], we create *elastic-bands* also for the humans and extend their hyper-graph structure to handle humans separately from ordinary obstacles using well-grounded *human-aware planning constraints*. The resulting scheme combines in one step the *robot-plans*, *human-plans* and *robot-reacts* process.

### *3.1 Elastic Band and Graph Optimization*

The timed elastic band approach [28] augments each of the $n \in \mathbb{N}$ 2D poses, defined as $\mathcal{P} := \{p_i = [x_i, y_i, \theta_i]^T\}_{i=0}^{n}$, of the robot path with time interval between each consecutive poses, denoted as $\mathcal{T} := \{\Delta t_j\}_{j=0}^{n-1}$. The resulting tuple $\mathcal{B} := \{\mathcal{P}, \mathcal{T}\}$ represents a trajectory that is subject to deformation by the optimization algorithm. In addition to the robot trajectory $\mathcal{B}_\mathcal{R}$, we also represent a set of human trajectories $\{\mathcal{B}_{\mathcal{H}_k}\}_{k=0}^{m}$ as timed elastic bands, when there are $m \in \mathbb{N}$ humans in the vicinity of the robot.

The gist of our approach is to jointly optimize the robot and human trajectories in terms of social and kinodynamic constraints. It requires to solve the following multivariate multi-objective optimization problem using the weighted-sum model:

$$f(\mathcal{B}_\mathcal{R}, \mathcal{B}_{\mathcal{H}_k}) = \sum_a \gamma_a f_a(\mathcal{B}_\mathcal{R}) + \sum_b \gamma_b f_b(\mathcal{B}_{\mathcal{H}_k}) + \sum_c \gamma_c f_c(\mathcal{B}_\mathcal{R}, \mathcal{B}_{\mathcal{H}_k}) \quad (1)$$

$$\{\mathcal{B}_\mathcal{R}^*, \mathcal{B}_{\mathcal{H}_k}^*\} = \underset{\{\mathcal{B}_\mathcal{R}, \mathcal{B}_{\mathcal{H}_k}\}}{\arg\min} f(\mathcal{B}_\mathcal{R}, \mathcal{B}_{\mathcal{H}_k}) \quad (2)$$

where $\{\mathcal{B}_\mathcal{R}^*, \mathcal{B}_{\mathcal{H}_k}^*\}$ denotes the set optimized robot and human trajectories. The component objective functions for the robot trajectory, human trajectories and human-robot social constraints are denoted by $f_a$, $f_b$ and $f_c$ respectively.

We inherit the kinodynamic and nonholonomic constraints from [28] imposed on the robot trajectory. These constraints enforce physical limits of velocity and acceleration between consecutive poses of the trajectory, a minimum clearance distance from obstacles for each of the poses, and fastest execution time for the whole trajectory. Similarly we impose kinodynamic constraints on human trajectories, with human velocity and acceleration limits obtained from empirical studies of pedestrians interaction data [31]. While predicting human trajectories the optimization algorithm tries to maintain nominal human velocity (and not the fastest velocity that human can move with). Constraints on human-robot cooperative motion are discussed in detail in Sec. 3.2.

In our approach we have modified the error function used for safety-clearance from obstacles as following,

$$f_{obs}(d, d_o, \varepsilon, S) = \begin{cases} \frac{(d_o+\varepsilon)-d}{Sd+1} & \text{if } 0 \leqslant d < (d_o+\varepsilon), \\ (d_o+\varepsilon)-d & \text{if } d < 0, \\ 0 & \text{otherwise.} \end{cases} \quad (3)$$

where $d$ is the distance between obstacle and a robot pose during current iteration of the optimization process, $d_o$ is lower bound for obstacle clearance, $\varepsilon$ is a parameter to control the accuracy of the approximation. This equation is non-linear between the lower bound and zero, and the parameter $S$ adjust the non-linearity. Such construction of the error function enables us to manipulate relative importance of safety clearance between robot-obstacle and robot-human. That is, we can make robot *push* itself more towards an obstacle rather than towards a human in constrained situations.

Similarly to [28], we have adopted the general optimization framework g$^2$o [32] which requires mapping of the least-squares problem into a graph representation. Each node in the graph represents a pose along the trajectory and edges that connect two nodes represent constraints, as shown in fig. 2. It is possible to write separate error functions for each of the optimization constraints. This graph based structure provides and easier interface to initialize and maintain the structure of variables and constraints that are used during the optimization process. The g$^2$o framework employs Levenberg-Marquardt to solve the non-linear least squares problem defined by the graph structure. The graph typically results in a sparse information matrix, consequently the g$^2$o framework uses state-of-the-art sparse linear system solvers [33]. Result of the optimization adjusts the position and orientation of each of the poses as well as time difference between the consecutive poses of the trajectory such that the whole trajectory minimizes the imposed constraints.

Since, the resulting trajectory is an optimally deformed version of the initial path, this representation of trajectory is similar to an *elastic band*. By adjusting weights on the constraints for robot and humans separately (the parameters $\gamma_a$ and $\gamma_b$ in equation 1), we can balance and tune the "tightness" of the elastic bands for effective effort sharing.

The optimization process runs in two computation loops. *Inner loop* corresponds to iterative loops for the least-squares solver. After each full run of the inner loop, the optimization process updates the graph structure using latest result of the solver. During this update, new nodes are added between two nodes in the graph if the time-difference between the two nodes exceeds certain threshold (which an optimization parameter) in order to maintain same time difference between each pair of neighbour nodes. Therefore, the human and robot time-difference nodes, shown as $h_i \Delta T_0$ and $r \Delta T_0$ in the graph structure are synchronized during this *outer loop* of graph update. Both inner and outer loops runs for several iterations, where the number of iterations directly affects the quality of produced trajectories and optimization time.

### 3.2 Social Constraints

We have selected the above mentioned graph-based solver because it enables us to introduce the social constraints and rules which are appropriate for efficient human-robot cooperative planning. Since we have the whole trajectories of human and robot at our disposal, we have added social constraints between human and robot nodes in the graph structure that correspond to the same time-step during their trajectories. That means, we add an edge for the *safety* constraint between $n^{\text{th}}$ nodes of human and robot trajectory, another edge between $n+1^{\text{th}}$ nodes, and so on. Nevertheless, we only add edges to the nodes corresponding to part of the human path that falls withing the *local planning* area centered at the robot.

*Safety Constraint:* The *safety* constraints uses proxemics based cost function to ensure minimum safety distance between corresponding human and robot poses. Therefore the error function associated with the safety constraint is,

$$f_{safety}(d, d_s, \varepsilon) = \begin{cases} (d_s + \varepsilon) - d & \text{if } d < (d_s + \varepsilon), \\ 0 & \text{otherwise.} \end{cases} \quad (4)$$

where $d_s$ is a lower bound on allowed safe distance between human and robot poses at the same time-stamp during the trajectory. There are multiple ways one can represent the footprint of the robot in this planning scheme, for simple circle to complex polygon. Humans footprint are represented as circle with specified radius. $d$ is the outer distance between human and robot. Calculating the outer distance between a polygon shaped robot and a circle shaped human is computationally more expensive than calculating outer distance between human and robot that are both circle shaped. Thus, it is advisable to represent the robot with a simpler footprint. That means, the optimization procedure ensures given safety distance between planned position of the robot and predicted position of the human at all future time points $t = 1, t = 2, \ldots$ up to the planning horizon.

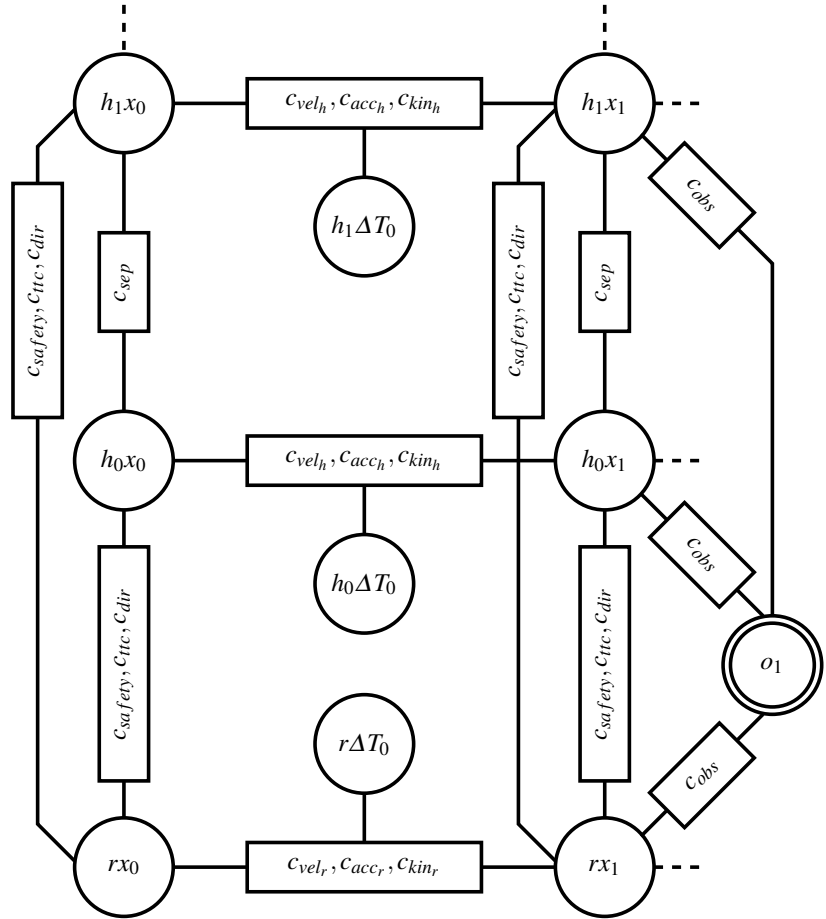

Fig. 2: Graph structure. The bottom row has consecutive nodes for robot trajectory ($rx_0$, $rx_1$, ...); edges enforcing velocity, acceleration, and kinodynamic constraints connects these nodes. We combined the three edges into one for easier depiction. Penalty imposed by these edges depends on time difference between consecutive nodes, therefore the *time-diff* node $r\Delta T_0$ connects with them. Pose and *time-diff* nodes are subject to change by the optimization process. Similarly, the middle and the top row represent trajectories for humans 0 and 1 respectively, albeit with different weights for constraints. Obstacles node ($o_1$), shown with double circle, is a fixed node, meaning the optimization process cannot alter its position. The edge $c_{obs}$ shows the constraint for keeping minimum distance from obstacles. The edge $c_{sep}$ represents the constraint to keep minimum separation between two humans. Nodes of the robot and a human that belong to the same time-step of their trajectories are connected by three edges ($c_{safety}$, $c_{ttc}$ and $c_{dir}$) that impose social constraints.

The graph structure injects an edge for this safety constraint not only to each human-robot trajectory pair but also to each human-human trajectory pair, albeit with different parameters for the error function. Such construction of the graph structure ensures that the optimization process respects proxemics cost between humans.

*Time-to-Collision Constraint:* A novel social constraint used in the proposed scheme is *time-to-collision*, that is, the projected time to a possible future collision with a human. Empirical studies have shown that *time-to-collision* between self and other governs the pedestrian interaction across wide variety of situation [34]. We make use of these results by applying higher cost to human-robot configurations that result in less *time-to-collision*. Our hypothesis is, the *time-to-collision* constraint will push the robot to act early enough, thus clearly showing the robot motion intentions to the human counterpart.

$$f_{ttc}(ttc, \tau, \varepsilon) = \begin{cases} \frac{[(\tau+\varepsilon)-ttc]\alpha}{C^2} & \text{if } ttc < (\tau+\varepsilon), \\ 0 & \text{otherwise.} \end{cases} \quad (5)$$

where $\tau$ is the lower bound on time at which the robot predicts a collision occurring with the human. In other words, $\tau = 8$ indicates that the robot start adding costs between the configuration (node) of human and robot whenever the computed time-to-collision goes below 8 s. The parameter $\alpha$ is a scaling parameter for strengthening or weakening the penalty on particular human robot configurations due to computed time to collision and $C^2$ is the squared distance between human and robot center points.

*ttc* is the predicted time to collision between a human and the robot. Here we consider both human and robot having a disc-like shape, however, with different radius. Time to collision defined as the time when the boundary of these two moving disc meet based on their current linear velocities. With this constraint our robot is able to proactively propose, a co-navigation solution sufficiently well ahead of time compared to other state-of-the-art approaches.

*Directional Constraint:* We have added the *directional* constraint, which defines a compatibility measure between human and robot configurations. Motivation for this social constraint comes from our previous work [15] which aims to improve the legibility of the robot motions. Similar to equation 4, the directional constraint is also lower bound by a threshold $\varsigma$, and defined as,

$$f_{dir}(c_{dir}, \varsigma, \varepsilon) = \begin{cases} (\varsigma + \varepsilon) - c_{dir} & \text{if } c_{dir} < (\varsigma + \varepsilon), \\ 0 & \text{otherwise.} \end{cases} \quad (6)$$

where the directional cost is defined as,

$$c_{dir} = \frac{\vec{v_\mathcal{R}} \cdot \overrightarrow{p_\mathcal{R} p_\mathcal{H}} + \vec{v_\mathcal{H}} \cdot \overrightarrow{p_\mathcal{H} p_\mathcal{R}}}{C^2} \quad (7)$$

$\vec{v_\mathcal{R}}$ and $\vec{v_\mathcal{H}}$ are robot and human velocity vectors respectively, and $\overrightarrow{p_\mathcal{R} p_\mathcal{H}}$ defines the vector from robot position to human position in 2D vector space and $C^2$ is the squared distance between center points of robot and human. This measure penalizes motions where human and robot are moving straight towards each other. Moreover, high relative velocity means higher penalty values. Directional constraint establishes tread-off between the effect of slowing down or changing the path.

With these social constraints we have tested the proposed planner in simulation and on two real robotic platforms. The Sec. 4 discusses the results of our tests in detail.

### 3.3 Human-Aware Planning Architecture

The elastic band approach can only locally deforms the robot trajectory, thus it requires an initial path to bootstrap the optimization. A simple grid-based *global* planning algorithm, for example $A^*$, is suitable for computing this initial path.

The well-known and versatile robot navigation architecture `move_base` [35] also differentiates between *global planning* for generating a path from the start position to the goal position in arbitrarily large environment, and *local planning* for avoiding immediate obstacles by locally modifying the robot trajectory. We have adopted `move_base` for our robot navigation architecture within which the proposed human-robot cooperative planner fits as a *local planner*. Fig. 3 depicts the full navigation architecture.

It is preferable to reason differently for moving and static humans in the robot operating environment. If position of any human is known at the time of global planning, it is better to incorporate the proxemics costs around static humans already while calculating the global path. Hence, we have developed a plugin for `layered_costmap` library [36] (which is part of the `move_base` framework) that add safety and visibility costs introduced by [14] around human positions in the occupancy grid-map.

We are using a separate module to predict first paths for tracked humans in the robot environment that the cooperative local planner uses for optimization. This allows us to switch between different prediction methods depending on the interactive situation at hand. For example, in a corridor crossing situation we use the heuristic that the most probable goal of the human is to go to other side of the corridor and so we create and imaginary goal position behind the current robot position when a human was first detected. In open areas, we use the velocity-obstacle method [37] to generate a short-term path for the tracked humans.

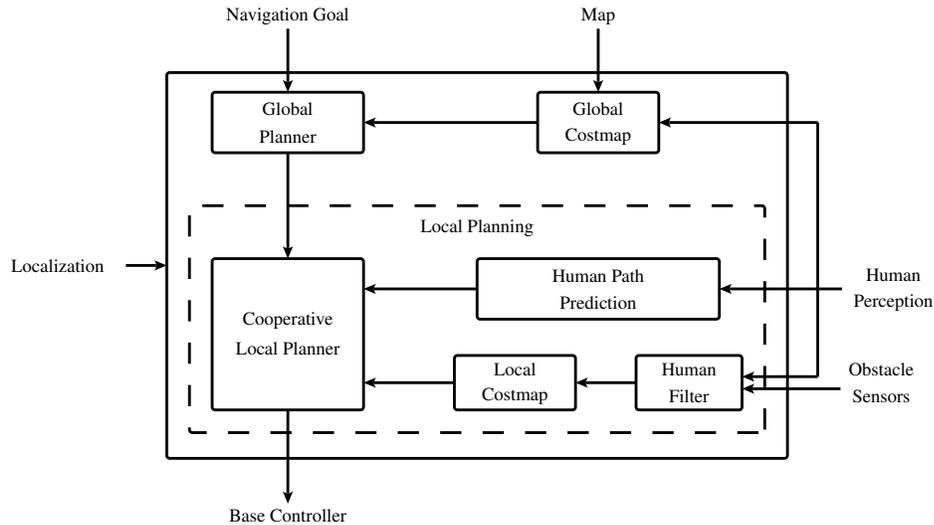

Fig. 3: Overview of the navigation software architecture.

Since in this cooperative planning scheme we are treating humans differently than other obstacles in the environment, we filter out the human position data from the obstacles detection sensors (laser scanner in out experiments).

## 4 EXPERIMENTS

Before conducting experiments in the real world, we tested the proposed social constraints and validated them in a simulation environment. We have designed a human navigation simulator with a framework similar to move_base, thereby using separate global and local planning modules. The global planning module uses the same global costmap that of the robot, for planning global paths for humans between given start and goal positions. As the local planning module we have developed simple teleport controller to update human position and velocity, making human move on the global path with a constant velocity. By exposing the human trajectories from the cooperative planner to the human navigation simulator as new paths for the humans to follow, we can simulate full interactive navigation situations. The human simulator can simulate motion of multiple humans, thus allowing us to test our planners also on semi-crowded environments. For simulating a PR2[1] robot, we have used the generic open source simulation engine MORSE[2]. Fig. 4 shows a screenshot of the simulation interface. In out previous work [38], we have compared the proposed cooperative planner with two other state-of-the-art human-aware navigation planners in several canonical path crossing situations.

---

[1] http://wiki.ros.org/Robots/PR2

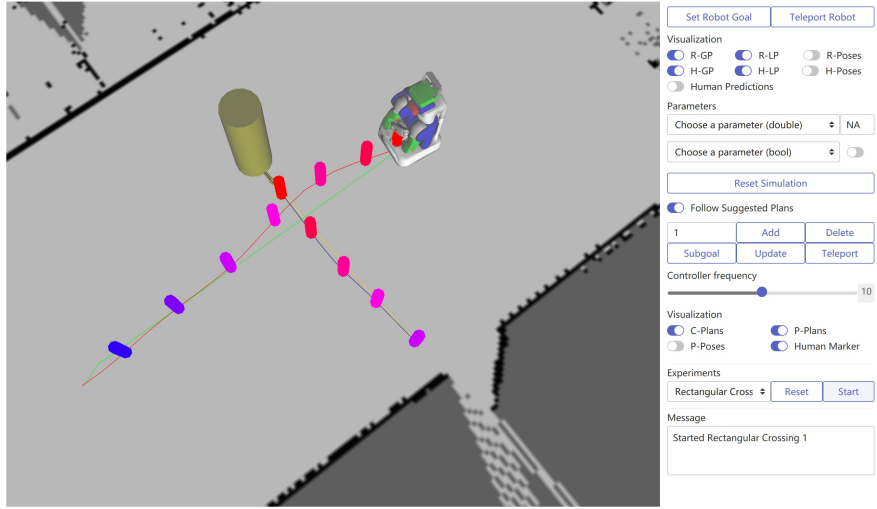

Fig. 4: Interface for testing the cooperative navigation planner in simulation.

We have ported the cooperative planner to two service robotic platforms, the PR2 and the Pepper[3] robot. For real world experiments, our objective is to validate the navigation planning system, not the human detection and tracking algorithms. Therefore we employed off-the-shelf motion capture system from OptiTrack[4]. It publishes positions and velocities of tracked humans at a certain frequency (10 Hz during our experiments[5]).

On the PR2 robot we have also activated a module for coordinating the head motion with the navigation planner to facilitate communication of robot's navigational intentions [39].

Fig. 5 demonstrates the capabilities of cooperative planning with a series of experiments in real world interactive situations. We have observed it performing better compared to purely reactive planning schemes. It should be noted that, for all interactive situations shown here the planer was not particularly "informed" about the task. The behavior such as stopping near the door and facilitating human in confined corridors emerged because of the integrated social constraints in the optimiza-tion framework. Fig. 6 show the effect of tuning the effort between a human and a robot for a shared navigation task.

---

[2] https://www.openrobots.org/wiki/morse

[3] https://www.ald.softbankrobotics.com/en/pepper

[4] http://www.optitrack.com/

[5] Although the motion capture system delivers data at higher frequency (about 100 Hz), we apply a moving average filter and re-sample the filtered data at 10 Hz to have better estimate of velocities

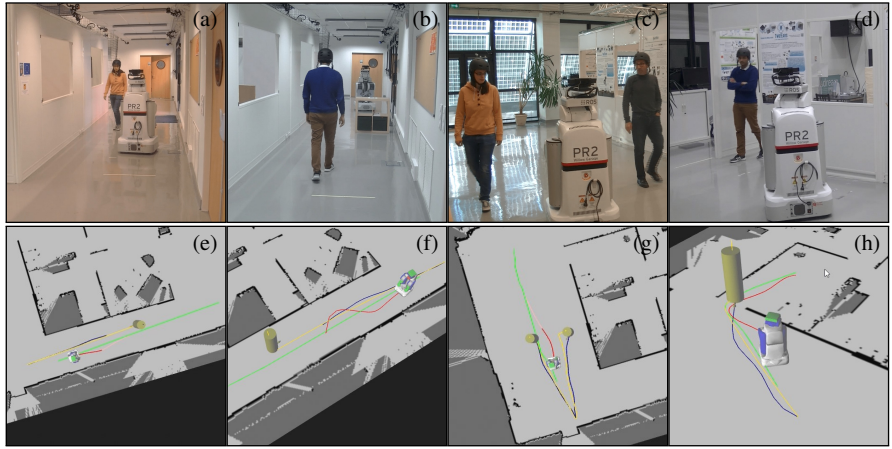

Fig. 5: Trajectories generated by cooperative planner for different interactive navigation tasks. (a and e) A corridor crossing situation where human and robot share effort to avoid colliding with each other. (b and f) A more confined corridor crossing situation where the robot facilitates the human to cross the corridor with sufficient space. (c and g) An open area crossing situation where two human decides to move on either side of the robot. Although in this situation the original plan suggest by the robot was to pass both persons on its right side, when the persons decide otherwise the robot complied and quickly adapted its own path accordingly. (d and h) A door crossing situation where the human wants to pass through a door, the robot facilitates the human by stopping near the door. The robot stops not because of a planning failure but because it has planned a cooperative strategy where it waits until the human passes through.

During a navigation task, if the human decides to move on other path than one suggested by the robot (e.g., choosing to pass by another side of the robot), the robot quickly adapts its trajectory. In situations where robot has enough space to move well advance in time, the robot proactively chooses a path that is both legible and comfortable for the human counterpart.

## 5 CONCLUSION AND FUTURE WORK

Proposed cooperative planning scheme has several advantages over state-of-the-art human-aware planning schemes. Our robot does not stay purely reactive but now it can also propose a path for the human assuming she will consider the proposed solution that benefits both agents. This is crucial especially in confined spaces, such as corridors where two agents can navigate only in side-by-side configuration.

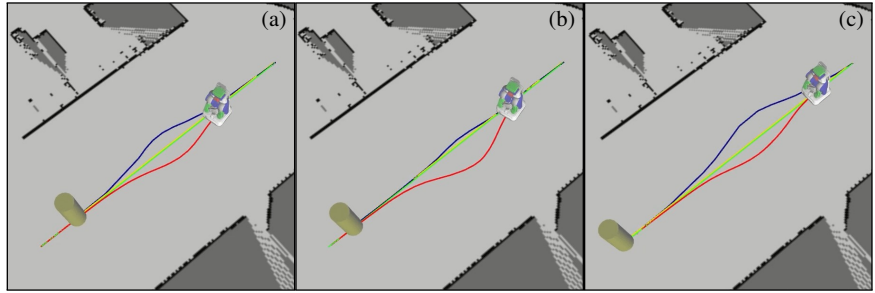

Fig. 6: Balancing the shared effort between human and robot. (a) Human and robot share equal effort. (b) The robot shares most of the effort and moves far away from predicted human path. (c) Although never used in real world situations, it is possible to design a rude behavior of the robot where it expects human to make more effort for avoiding a collision.

This approach opens up a new avenue for quickly testing and comparing different social constraints. We plan to define new social constraint that will enable further interactive motion scenarios such as actively approaching a person, queuing behavior in long hallway like environments with multiple humans, guiding a group of people, and more. We are preparing for a series of experiments and a user study which compares the cooperative planner against other proactive planning approaches.

The computational cost of the cooperative planner increases with the number of humans surrounding the robot as well as number and type of social constraints used for optimization. This is a limiting factor for using the cooperative planner in crowded situation. We plan to improve over this limitation by parallelly evaluating the error functions and utilizing GPU capabilities provided by state-of-the-art sparse matrix solvers.

**Acknowledgements** This work is supported by the European Union's Horizon 2020 research and innovation programme under grant agreement No. 688147 (MuMMER project).